\documentclass{article}

\usepackage[preprint]{template/iclr2027_conference}
\usepackage{times}
\usepackage[T1]{fontenc}
\usepackage[utf8]{inputenc}
\usepackage{microtype}
\usepackage{graphicx}
\usepackage{xcolor}
\usepackage{placeins}
\usepackage{subcaption}
\usepackage{booktabs}
\usepackage{algorithm}
\usepackage{algorithmic}
\usepackage{amsmath}
\usepackage{amssymb}
\usepackage{mathtools}
\usepackage{amsthm}
\usepackage{hyperref}
\hypersetup{
  hidelinks,
  pdftitle={Probabilistic State Space Enumeration in Neural Network Computation},
  pdfsubject={Preprint},
  pdfkeywords={probabilistic state space enumeration, neural networks, structured pruning, language models}
}
\usepackage{xurl}
\usepackage[capitalize,noabbrev]{cleveref}

\theoremstyle{plain}
\newtheorem{theorem}{Theorem}[section]
\newtheorem{proposition}[theorem]{Proposition}

\theoremstyle{definition}

\theoremstyle{remark}

\title{PROBABILISTIC STATE SPACE ENUMERATION IN NEURAL NETWORK COMPUTATION}

\newcommand{\authoremail}[1]{{\normalfont\small\sffamily\href{mailto:#1}{#1}}}
\author{%
  \begin{tabular}[t]{@{}*{3}{p{\dimexpr(\textwidth-2\tabcolsep)/3\relax}@{}}}
    \textbf{Victor OK Li}\textsuperscript{1,2,*} &
    \textbf{Yang Han}\textsuperscript{1,3} &
    \textbf{Jacqueline CK Lam}\textsuperscript{1,*} \\
    \authoremail{vli@elohax.ai} & \authoremail{yhan@elohax.ai} & \authoremail{lajacob@elohax.ai} \\
    \authoremail{vli@hkust-gz.edu.cn} & \authoremail{yanghan001@cuhk.edu.hk} & \\
  \end{tabular} \\
  \noalign{\vskip 11pt}
  \normalfont \textsuperscript{1}ElohaX Ltd. \\
  \normalfont \textsuperscript{2}The Hong Kong University of Science and Technology (Guangzhou), China \\
  \normalfont \textsuperscript{3}The Chinese University of Hong Kong, Hong Kong SAR, China \\[2pt]
  \normalfont \textsuperscript{*}Corresponding authors%
}
\hypersetup{pdfauthor={Victor O. K. Li; Yang Han; Jacqueline C. K. Lam}}

\begin{document}
\maketitle

\begin{abstract}
Can we faithfully approximate the output of a dense Transformer multilayer perceptron
(MLP) by enumerating only a small set of probable channel configurations?
Our primary objective is to characterize and control approximation error under a
fixed mask budget. Probabilistic state space enumeration (PSSE) links the selected
masks to the residual in the MLP output. For a partition into channel groups,
probability weighting makes the expectation over all possible masks equal to
the dense output. We derive the exact residual for any retained set by weighting each
group's output by the fraction of its inclusion probability outside the set. The
omitted probability mass multiplied by the largest masked-output norm therefore
bounds the absolute error. To select masks, NORDER-II adapts ORDER-II from network
reliability. Under independent group inclusions, it computes an ordering once per
layer from frozen weights and finds the $B$ masks with greatest total probability
mass without listing all possibilities. With 32 groups, the mean number of masks
needed to capture 95\% of the mass ranges from 1,020 to 4,725 across models, out of
more than four billion possibilities. Across 168 MLP layers in six language models,
this approximation yields lower mean relative error than both ordinary and
bias-corrected uniform sampling in all 1,008 equal-count comparisons. Our secondary
objective is to improve inference efficiency through pruning. We count the distinct
channels covered by each layer's retained masks and use those counts to allocate a
fixed projection budget offline. The pruned models achieve lower held-out perplexity
than both top-$k$ baselines in all 36 settings at matched MLP projection FLOPs.
Target projection savings of 10\%, 20\%, and 30\% yield mean decoder MLP speedups of
$1.11\times$, $1.23\times$, and $1.36\times$ over dense MLPs, respectively.
\end{abstract}

\section{Introduction}
\label{sec:intro}

In large language models, each Transformer multilayer perceptron (MLP) sums
contributions from thousands of intermediate channels for every token.
These wide projections can account for a substantial share of inference time~\citep{liu2023dejavu}.
One way to reduce this cost is to rank channels individually and retain a single
subset, for example by scoring each channel from its outgoing weights~\citep{stahlberg2025outgoing}.
Pruning channels saves work, but the error depends on how the omitted channel
contributions add together. Even small contributions can reinforce one another.
Can we approximate the dense output by combining several channel subsets, with a
criterion for deciding when to stop?

In network reliability analysis, one can estimate the probability that a network
remains connected by visiting the most likely configurations of working and failed
links first~\citep{li1984performance,lam1986improved}. We bring this idea to
Transformer MLPs through \emph{probabilistic state space enumeration} (PSSE).
We partition the channels into $G$ groups and map a link's up/down status to the
inclusion or exclusion of a group. Each subset is a joint state. We derive each
group's inclusion probability from the mean of its channels' weight-entropy scores.
Dividing each included group's contribution by its inclusion probability makes
the expected state output equal to the dense output (Theorem~\ref{thm:representation-equivalence}).
Our adaptation of ORDER-II, NORDER-II, visits states in decreasing probability until
their cumulative mass reaches a target $\tau$. For this \emph{prefix}, the
approximation error is at most the omitted probability mass times the maximum
norm of a state output (Theorem~\ref{thm:mass-truncation}).

Across six language models from the Llama, Gemma, and Qwen families and two
datasets, the 95\%-mass prefix yields lower mean relative error than sampling the same
number of states uniformly at random, with or without bias correction. This holds for
each of the 168 MLP layers at all three group counts, for 1,008 comparisons in total.
At all three group counts, PSSE also has lower median and 99th-percentile relative
error than a single top-$k$ mask, with $k$ set to the rounded mean number of active
groups per selected state in each layer (Figure~\ref{fig:local-ordering}b).
For downstream inference, we prune each model once per budget,
using the same validation-selected group count for PSSE and both top-$k$ baselines. Across six models and two datasets,
PSSE gives lower held-out perplexity than both baselines at matched projection FLOPs
in all 12 comparisons at each of the three MLP projection savings targets
(10\%, 20\%, and 30\%) (Table~\ref{tab:main-ppl}). At these targets,
PSSE's pruned decoder MLPs achieve mean speedups of
$1.11\times$, $1.23\times$, and $1.36\times$ over dense MLPs (Table~\ref{tab:mlp-latency}).

We make four contributions.
\begin{itemize}
    \item \textbf{Mass optimality at a fixed state budget.} Under independent group
    inclusions, NORDER-II returns the $B$ most probable joint states. No other set of
    $B$ states captures more mass. For a common bound on masked-output norms, this
    prefix therefore minimizes the omitted-mass error bound (Theorem~\ref{thm:mass-truncation}).
    \item \textbf{Exact MLP truncation residual.} Each group coefficient $a_g$ in
    Eq.~\eqref{eq:coefficient-collapsed-prefix} is the conditional probability that a
    state is retained given that group $g$ is included. Proposition~\ref{prop:exact-truncation-error}
    gives the exact residual $y(x)-\hat y_{\mathcal{P}}(x)=\sum_{g=1}^{G}(1-a_g)c_g(x)$. Because these vectors
    can cancel, the set that captures the most mass need not minimize actual output error.
    \item \textbf{Local approximation.} We isolate the effect of state ordering by
    comparing estimators at equal state counts on identical layer inputs. With group
    probabilities held fixed, raising the mass target from 95\% to 99\% lowers mean error
    in all 24 combinations of model, dataset, and group count (Appendix~\ref{sec:sensitivity}).
    \item \textbf{Downstream inference.} We use each layer's prefix to allocate
    a fixed MLP projection budget and use one channel-or-group rule per layer across budgets.
    At matched MLP projection FLOPs and a shared validation-selected group count per model,
    PSSE achieves lower held-out perplexity than both top-$k$ baselines in all 36
    settings, with a median reduction of 15.6\% relative to the better baseline
    (Table~\ref{tab:main-ppl}).
    In a width-allocation ablation at matched projection FLOPs, allocating widths by
    the distinct channels in each layer's prefix rather than by its expected active-channel
    count lowered held-out perplexity in 31 of 36 settings, with a median reduction of
    11.7\% across all 36 (Appendix Table~\ref{tab:width-allocation}).
    Its decoder MLPs also have lower summed latency than dense MLPs in all 36 settings.
\end{itemize}

\section{Related Work}
\label{sec:related_work}

\paragraph{Sparse and efficient neural inference.} Traditional approaches include unstructured weight pruning \citep{han2015learning} and structured pruning of channels, filters, or units to improve hardware efficiency \citep{he2017channel}. In transformers, prior work has shown that many attention heads are redundant and can be removed with little impact on performance \citep{michel2019sixteen,voita2019analyzing}. PSSE approximates the dense MLP output by summing the most probable states' contributions. Layers with more concentrated distributions reach the same mass target with fewer states.

\paragraph{Conditional computation and mixture models.} Conditional computation methods, such as mixture-of-experts models, dynamically route inputs to subsets of parameters \citep{shazeer2017outrageously}. A mixture-of-experts layer learns a routing function that selects experts for each input. PSSE derives a distribution over subsets of channels from the frozen weights of a pretrained MLP. Its state ordering is fixed across inputs and requires no learned router.

\paragraph{Discrete subset sampling.} SIMPLE studies differentiable fixed-cardinality subset sampling during learning, using discrete $k$-subsets in the forward pass and exact marginals for gradients \citep{ahmed2023simple}. Our framework instead uses complete, variable-cardinality Bernoulli configurations; NORDER-II orders them to a mass target or state budget, without gradients or a cardinality constraint. Appendix~\ref{sec:extended-related-work} gives the classical enumeration background and further comparisons with input-dependent sparsity, sparse attention, and analyses of effective model usage.

\section{Methodology}
\label{sec:method}

\subsection{Problem Setting and Probabilistic State Spaces}
\label{sec:problem}

We construct an \emph{auxiliary probabilistic state space} for a neural module.
Given an input or intermediate activation $x$, let $\mathcal{S}$ be a finite set of
deterministic masked computations. Each state $s\in\mathcal{S}$ specifies which
components contribute to the output $f(x,s)$. The masks define alternative computations
for the same pretrained module. We associate them with a probability mass
$P_\theta(s \mid x)$, $s \in \mathcal{S}$,
satisfying $\sum_{s \in \mathcal{S}} P_\theta(s \mid x) = 1$.
In our main experiments, each MLP layer uses the same distribution for every input.

The state computation and probability model are constructed so that the dense module
output has the exact representation
\begin{equation}
\label{eq:dense}
y(x)
\;=\;
\sum_{s \in \mathcal{S}} P_\theta(s \mid x)\, f(x,s).
\end{equation}
Our MLP construction uses complete binary configurations over channel groups, with
heuristic Bernoulli salience scores and importance-corrected state computations. The
inclusion probabilities determine traversal order and finite-prefix quality, while the
importance correction preserves the exact full expectation.
Appendix~\ref{sec:controlled-illustrations} provides controlled illustrations of
this abstraction, including an exact comparison between NORDER-II and exhaustive
enumeration and a trained one-hidden-layer network with group-level binary states.

\subsection{Transformer MLP Instantiation}
\label{sec:transformer-mlp}

For input $x$, the gate and up projections of a gated Transformer MLP produce the
intermediate activation
\begin{equation}
h(x)=\phi(W_{\mathrm{gate}}x)\odot W_{\mathrm{up}}x.
\end{equation}
Here $\phi$ is the gate activation and $\odot$ denotes elementwise multiplication.
Within each block, we partition the intermediate coordinates into $G$ nearly equal
contiguous sets $I_g$. Each set indexes matching rows of $W_{\mathrm{gate}}$ and
$W_{\mathrm{up}}$ and columns of $W_{\mathrm{down}}$. Its contribution is
\begin{equation}
c_g(x)=W_{\mathrm{down}}[:,I_g]h_{I_g}(x),\qquad
y(x)=\sum_{g=1}^G c_g(x).
\end{equation}

If group $g$ is included with probability $\pi_g$, its expected contribution
under simple masking is $\pi_g c_g(x)$. Dividing by $\pi_g$ when the group is
included restores its full weight. For a binary mask $z$ with independent entries $z_g\sim\mathrm{Bernoulli}(\pi_g)$,
the importance-corrected masked output is
\begin{equation}
\label{eq:local-estimator}
F(x,z)=\sum_{g=1}^G \frac{z_g}{\pi_g}c_g(x),
\end{equation}
where the independent joint-state probability is
\begin{equation}
\label{eq:independent-joint-prob}
P(z\mid x)=\prod_{g=1}^G \pi_g^{z_g}(1-\pi_g)^{1-z_g}.
\end{equation}

\begin{theorem}[Representation equivalence]
\label{thm:representation-equivalence}
For any $\pi_g\in(0,1]$, the auxiliary state computation in
Eq.~\eqref{eq:local-estimator} satisfies $\mathbb{E}_z[F(x,z)]=y(x)$.
\end{theorem}
\begin{proof}
By linearity of expectation and $\mathbb{E}[z_g]=\pi_g$,
$\mathbb{E}_z[F(x,z)]
=\sum_{g=1}^G\frac{\mathbb{E}[z_g]}{\pi_g}c_g(x)
=\sum_{g=1}^G c_g(x)=y(x)$.
\end{proof}

Following \citet{stahlberg2025outgoing}, we score each intermediate channel by
the concentration of its outgoing weights. For a nonzero column $j$ of
$W=W_{\mathrm{down}}$ with $d$ output coordinates, set
$q_{ij}=|W_{ij}|/\sum_{k=1}^d|W_{kj}|$. The channel's weight-entropy score is the KL divergence
from uniform, $s_j=\log d-H(q_{\cdot j})=\sum_{i:q_{ij}>0}q_{ij}\log(dq_{ij})$,
where $H$ is Shannon entropy. We set $s_j=0$ for a zero column.
A higher score means the weight magnitude is concentrated on fewer output coordinates.
The group score $u_g$ and its inclusion probability are
\begin{equation}
\label{eq:mlp-prob-model}
u_g=\frac{1}{|I_g|}\sum_{j\in I_g}s_j,\qquad
\tilde u_g=\frac{u_g-\mu_u}{\max(\sigma_u,\epsilon)},\qquad
\pi_g=\pi_{\min}+(1-2\pi_{\min})\,\mathrm{sigmoid}(\alpha\tilde u_g+\beta).
\end{equation}
Here $\mu_u$ and $\sigma_u$ are the mean and population standard deviation of the
$G$ group scores in the layer. In the main experiments $\alpha=8$, $\beta=0$, $\pi_{\min}=0$, and $\epsilon=10^{-8}$.
A larger slope sharpens the group probabilities. A group at the mean score has
inclusion probability $0.5$. The sigmoid gives positive probabilities with zero
floor. We bound probabilities away from zero and one for numerical stability
(Appendix~\ref{sec:sensitivity}).
These heuristic probabilities depend only on the frozen weights and the channel partition,
so we compute them before evaluating any group contribution $c_g(x)$.
Table~\ref{tab:scorer-comparison} compares the fixed score with the token-specific
group score $\|c_g(x)\|_2$ across all six models.
The contribution-norm score requires computing every group vector for the current token, and those
vectors already sum to the dense MLP output.

Theorem~\ref{thm:representation-equivalence} holds for any grouping and any positive
inclusion probabilities. These choices determine which states appear first and how
accurately a finite prefix approximates the dense output. Appendix~\ref{sec:ffn-example}
illustrates the identity for three groups and all eight masks.

\subsection{Probability-Mass Truncation}
\label{sec:mass-truncation}

The approximation restricts Eq.~\eqref{eq:dense} to a probability-ordered prefix.

For a mass target $\tau \in (0,1]$, the truncation seeks the shortest
probability-ordered prefix $\widehat{\mathcal{S}} \subset \mathcal{S}$ such that
\begin{equation}
\sum_{s \in \widehat{\mathcal{S}}} P_\theta(s \mid x) \;\ge\; \tau.
\end{equation}
The dense expectation in Eq.~\eqref{eq:dense} is then approximated by
\begin{equation}
\label{eq:mass-trunc}
\widehat{y}(x)
\;=\;
\sum_{s \in \widehat{\mathcal{S}}} P_\theta(s \mid x)\, f(x,s).
\end{equation}

\begin{theorem}[Mass truncation bound]
\label{thm:mass-truncation}
Assume that $\|f(x,s)\| \le B(x)$ for all $s \in \mathcal{S}$.
Let $\widehat{\mathcal{S}} \subset \mathcal{S}$ satisfy
$\sum_{s \in \widehat{\mathcal{S}}} P_\theta(s \mid x) \ge \tau$.
Then the approximation in Eq.~\eqref{eq:mass-trunc} obeys
\begin{equation}
\|y(x) - \widehat{y}(x)\|
\;\le\;
(1-\tau)\, B(x).
\end{equation}
\end{theorem}
The proof is given in Appendix~\ref{sec:bound}.

At a fixed mass target, the prefix length measures how many states account
for that share of the distribution. The approximation combines the corresponding
masked outputs using the pretrained weights.

For a probability-ordered prefix
$\mathcal{P}$, NORDER-II uses
\begin{equation}
\label{eq:prefix-estimator}
\hat y_{\mathcal{P}}(x)
=\sum_{z\in\mathcal{P}}P(z\mid x)F(x,z).
\end{equation}
We can evaluate this sum over states using one weight per group. Substituting
Eq.~\eqref{eq:local-estimator} into Eq.~\eqref{eq:prefix-estimator} and exchanging
the finite sums gives
\begin{equation}
\label{eq:coefficient-collapsed-prefix}
\begin{aligned}
\hat y_{\mathcal{P}}(x)
&=\sum_{g=1}^G a_g(\mathcal{P},x)c_g(x),\\
a_g(\mathcal{P},x)
&=\frac{1}{\pi_g}\sum_{z\in\mathcal{P}}P(z\mid x)z_g .
\end{aligned}
\end{equation}
The coefficient $a_g$ is the total probability of retained states that include
group $g$, divided by $\pi_g$. This lets us weight each group contribution once. After the group
contributions $\{c_g(x)\}_{g=1}^G$ are available, aggregation costs
$O(|\mathcal{P}|G+Gd)$ with a dense scan over a prefix of length $|\mathcal{P}|$ and
$O(A(\mathcal{P})+Gd)$ with sparse active-state accumulation, where
$A(\mathcal{P})=\sum_{z\in\mathcal{P}}\sum_g z_g$ is the number of active group
occurrences in the prefix and $d$ is the output dimension. This avoids
constructing a separate $d$-dimensional masked output for every retained state.
For a fixed weight-entropy prefix, each $a_g$ is constant across tokens.
Scaling the columns $W_{\mathrm{down}}[:,I_g]$ by $a_g$ once implements the
entire prefix as a single down projection.
The full state space recovers $y(x)$ exactly; for a truncated prefix, omitted mass
controls the worst-case error under the bounded-output condition in
Theorem~\ref{thm:mass-truncation}.

\subsection{Joint Binary Neural State Spaces}
\label{sec:joint-states}

For a given input $x$, consider $G$ conditionally independent binary factors
$z=(z_1,\ldots,z_G)$, where $z_g\in\{0,1\}$. In the MLP experiments, each factor
indicates whether a group of intermediate channels is included. The same enumeration
applies to binary choices over other components, such as attention heads or blocks.

For each factor, let
$\pi_g = P(z_g=1\mid x)$.
The base state takes the more probable value of each factor, with $b_g=1$
when $\pi_g\geq 0.5$ and $b_g=0$ otherwise.
Any joint state can be represented as a flip subset $S\subseteq\{1,\ldots,G\}$:
factors in $S$ are flipped away from their base values, and all other factors remain
at the base mode. Define
$p^{\mathrm{base}}_g = \max(\pi_g,1-\pi_g), \quad p^{\mathrm{alt}}_g = \min(\pi_g,1-\pi_g)$,
and
$r_g = \frac{p^{\mathrm{alt}}_g}{p^{\mathrm{base}}_g}, \qquad \lambda_g = -\log r_g$.
The base state has no flips. A flip subset $S$ has probability
\begin{equation}
\label{eq:joint-state-prob}
P(S\mid x)
= P(\mathrm{base}\mid x)\exp\!\left(-\sum_{g\in S}\lambda_g\right).
\end{equation}
Thus ordering joint states by probability is equivalent to ordering flip subsets by
the additive penalty $\sum_{g\in S}\lambda_g$.

\subsection{NORDER-II Enumeration}
\label{sec:norder2}

ORDER-II ranks configurations of working and failed components in network
reliability~\citep{li1984performance,lam1986improved}. NORDER-II maps these
two states to including or excluding a group of neurons; Eq.~\eqref{eq:local-estimator}
defines the masked output for each joint state. After sorting the flip penalties, the empty flip
set generates the singleton with smallest penalty. A nonempty set with largest sorted
index $j<G$ generates two children by replacing $j$ with $j+1$ or appending $j+1$. A priority queue
returns the available set with smallest total penalty. The first $B$ states therefore
capture the greatest possible mass for that count, with ties interchangeable.

\begin{algorithm}[!htbp]
\caption{NORDER-II for Joint Binary Neural States}
\label{alg:norder2}
\begin{algorithmic}[1]
\REQUIRE Factor probabilities $\pi_1,\ldots,\pi_G$, mass target $\tau$, state budget $B$
\ENSURE Probability-ordered joint-state prefix
\STATE Compute base modes $b_g$, penalties $\lambda_g$, and sorted order of factors
\STATE Initialize prefix $\mathcal{P}\leftarrow \emptyset$ and cumulative mass $\rho\leftarrow 0$
\STATE Initialize heap with the empty flip subset $S=\emptyset$
\WHILE{heap is nonempty and $\rho<\tau$ and $|\mathcal{P}|<B$}
    \STATE Pop the subset $S$ with smallest penalty
    \STATE Convert $S$ to its binary state and append it to $\mathcal{P}$
    \STATE Add its probability from Eq.~\eqref{eq:joint-state-prob} to $\rho$
    \STATE Push valid ORDER-II children of $S$ that have not been visited
\ENDWHILE
\STATE \textbf{return} $\mathcal{P}$
\end{algorithmic}
\end{algorithm}

Sorting the $G$ flip penalties costs $O(G\log G)$. Each
emitted state requires at most one heap removal and two insertions, for a total of
$O(B\log B)$ comparisons of scalar priority keys. Writing and storing
explicit state masks require $O(BG)$ time and space. These enumeration
costs exclude scoring the weights and computing group contributions. With the
weight-entropy scorer, we can compute the group probabilities and state ordering
once per layer and reuse them for every token.

{
\subsection{From state prefixes to pruned MLPs}
\label{sec:prefix-packing}
}
Before inference, we score channels from the frozen down-projection weights and
use the resulting state prefixes (Algorithm~\ref{alg:norder2}) to distribute a fixed
MLP projection budget across layers. Let $n_\ell$ count the
distinct channels included in at least one state of layer $\ell$'s prefix. Given a
total retained width $W$, layer $\ell$ receives a target of $t_\ell=W n_\ell/\sum_j n_j$
channels. To allow selection of whole groups at the same width, we rank groups by
decreasing inclusion probability $\pi_g$. Each feasible width is the total size of
the first $k$ groups for some $k\geq 1$. We choose feasible widths $w_\ell$ that
minimize $\sum_\ell(w_\ell-t_\ell)^2$ subject to $\sum_\ell w_\ell=W$.

At a given width, a layer can keep either the individual channels with the
highest weight-entropy scores $s_j$ or the highest-ranked whole groups.
For each layer, we evaluate both rules at the width assigned by each budget.
We average each rule's next-token validation loss across budgets with equal weight
and use the rule with the lower mean at every budget. We then evaluate the model
with all MLPs pruned on a separate validation block and choose one group count
for the model across all budgets.
After validation, we prune each MLP once per budget. For each discarded channel,
we remove one row from each of the gate and up projection matrices and the matching
column from the down projection matrix. Every token then uses the same smaller dense
matrices with the retained weights unchanged. The MLP never computes the discarded
channels, so each forward pass requires fewer multiplications. No state enumeration
or channel selection takes place during inference.

\section{Experimental Setup}
\label{sec:exp}

\subsection{Models and data}
\label{sec:exp:model-data}
We evaluate six pretrained language models, Llama-3.2-1B and
-3B~\citep{meta2024llama32}, Gemma-3-1B-pt and -4B-pt~\citep{gemma2025technical},
and Qwen3-1.7B-Base and -8B-Base~\citep{yang2025qwen3}. They have 16, 28, 26, 34,
28, and 36 text-decoder layers, respectively. We use bfloat16 (BF16), a 16-bit
floating-point format, for all model weights, including those in the pruned MLPs.
We use WikiText-2~\citep{merity2016pointer}
and biomedical abstracts from the \href{https://huggingface.co/datasets/MedRAG/pubmed/tree/33da3593d5756bc04c8909f170003c0b14197957}{MedRAG PubMed corpus}~\citep{xiong2024benchmarking}. Within each
perplexity comparison, all methods use the same model, tokenizer, and held-out tokens.

To measure how closely each method approximates a dense MLP, we use 256 prompts
per dataset, each limited to 128 tokens. We compare layer outputs at the last
non-padding token of each prompt, giving every method the same input from the dense
model. For WikiText-2, we sample validation passages uniformly without replacement.
For PubMed, we use a deduplicated subset with disjoint article sets for local
approximation and downstream evaluation.

For downstream inference, we consider $G\in\{8,12,16,20,24,28,32\}$.
For each candidate group count, we use 4,096 WikiText-2 validation targets to
choose between retaining individual channels and retaining whole groups in each
layer, keeping the other layers dense. We then prune all layers
and score a separate block of 4,096 WikiText-2 validation targets. We select
the $G$ that minimizes PSSE's mean next-token negative log-likelihood, giving
the 10\%, 20\%, and 30\% savings targets equal weight. PSSE and budget-matched
top-$k$ baselines (see Section~\ref{sec:packed}) use this same $G$ at all three budgets on both test corpora.
We score 8,192 held-out targets per model and dataset in 1,024-token windows
with stride 512 (Appendix~\ref{sec:packed-protocol}).

\subsection{Local approximation baselines}
\label{sec:exp:baselines}
We divide each layer's channels into $G\in\{8,16,32\}$ nearly equal contiguous groups.
NORDER-II selects states until their mass reaches 95\%, with no state-count cap.
Both random baselines match this prefix length $m$. For each token, we sample $m$ distinct
states uniformly from the $2^G$ possibilities and sum their $P(z)F(x,z)$ terms to obtain $R$.
Each state is sampled with probability $m/2^G$, so the Horvitz--Thompson (HT) estimate
$(2^G/m)R$ is unbiased~\citep{horvitz1952}. For the single top-$k$ mask, we set $k$ to
the rounded mean number of active groups per state in that layer's prefix. The mask sums
the original contributions $c_g(x)$ of the $k$ groups with the highest inclusion probabilities.
We measure relative error $\|\hat y-y\|_2/\|y\|_2$ against the dense output for each prompt
and average within each layer.

\subsection{Pruned MLPs at matched projection FLOPs}
\label{sec:packed}
We target savings of 10\%, 20\%, and 30\% relative to the dense MLP projections.
For a model with hidden size $h$, the gate, up and down projections together cost
$6hW$ FLOPs per token across all layers, counting a multiply-add as two FLOPs.
We form contiguous groups whose widths are multiples of 128 channels and use the
largest total width $W$ feasible for all seven candidate group counts within each budget. All three methods
therefore have the same projection cost.

PSSE allocates widths from the 95\%-mass prefixes as described in
Section~\ref{sec:prefix-packing}. Both baselines use the same feasible widths and
allocation procedure but target $W/L$ channels in each of the $L$ layers.
Channel top-$k$ keeps the highest-scoring individual channels in each layer.
Grouped top-$k$ keeps the highest-ranked whole groups at the same layer
widths. All three methods use the original pretrained weights on the retained channels.

We measure decoder MLP latency on the same GPU for all methods. Each forward
pass processes one 512-token sequence with the key--value cache disabled. CUDA events
time each layer's MLP, including its projections, activation, and gating. We sum the
layer times within each pass and take the median over 40 passes per method and
setting. Weight scoring, rule selection, and pruning take place before timing.

\subsection{Computing infrastructure}
We implement the experiments in PyTorch and Hugging Face Transformers.
For downstream calibration, perplexity evaluation, and latency measurements,
we use one NVIDIA GeForce RTX~5090 GPU (32\,GB) per model.

\section{Results}
\label{sec:results}

\subsection{Probability ordering and local approximation error}
\begin{figure}[!htbp]
\centering
\includegraphics[width=\linewidth]{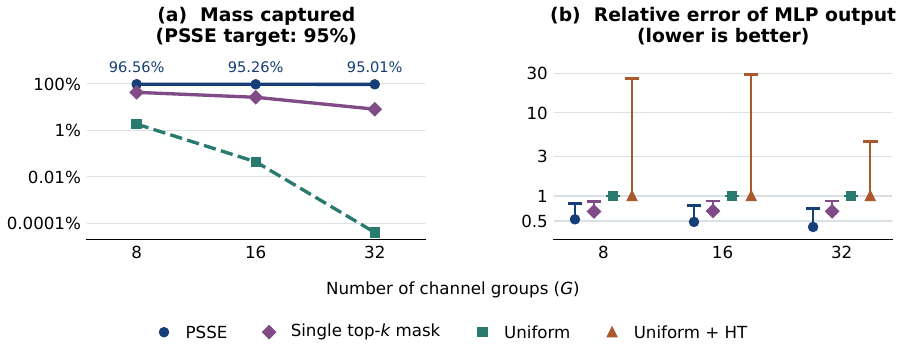}
\caption{Captured probability mass and relative MLP output error (logarithmic vertical axes). (a) PSSE stops when
its $m$ most probable states reach 95\% mass. A uniform sample of $m$ states captures
$m/2^G$ in expectation. For the single mask, the plotted mass is its joint probability
(Eq.~\eqref{eq:independent-joint-prob}). Points are medians of 12 model--corpus means.
(b) Markers show median error and caps show the 99th percentile across 86,016 prompt--layer
comparisons per $G$. The single mask sums contributions from the $k$ groups with the highest inclusion probabilities, with $k$ set
to the layer's rounded mean active count per selected state. Uniform and Uniform + HT use the same
draws (Appendix Table~\ref{tab:main-comparison}, panel B).}
\label{fig:local-ordering}
\end{figure}

PSSE's $m$ most probable states capture at least 95\% mass. A uniform sample of $m$ states
captures $m/2^G$ in expectation (Figure~\ref{fig:local-ordering}a). PSSE has lower mean
relative error than both random estimates in all 1,008 layer--corpus--$G$ comparisons.
At $G=8$, 16, and 32, its error is lower than both baselines' in 99.75\%, 99.92\%, and 100\% of the 86,016
prompt--layer comparisons per $G$, respectively. Increasing $G$ from 8 to 16 and then to 32
reduces mean PSSE error in all 12 model--corpus pairs at each step and requires longer prefixes.
At $G=8$, mean prefix lengths range from 3.7 to 5.9 states out of 256. Across all 168 layers,
selected states contain a mean of 3.84 of 8, 7.24 of 16, and 13.58 of 32 active groups.
The single top-$k$ mask captures median masses of 42.99\%, 26.32\%, and 8.08\%, respectively.
PSSE has lower mean relative error than the single top-$k$ mask in all 36
combinations of model, corpus, and group count (Appendix Table~\ref{tab:main-comparison}, panel A).
Its pooled median and 99th-percentile errors are also lower at each group count
(Figure~\ref{fig:local-ordering}b and Appendix Table~\ref{tab:main-comparison}, panel B).
At $G=8$, the uncorrected random sum has mean error 0.978--0.989, close to the relative error
of 1 for a zero estimate. Scaling by $2^G/m$ removes bias but amplifies errors on draws
containing a high-probability state. Both random estimates have median error near 1, but HT
has larger 99th-percentile error at every $G$ (Figure~\ref{fig:local-ordering}b).

\subsection{Sensitivity to group partition and probability settings}
We test sensitivity at $G=16$ and 32. With a 95\% mass target and probability floor
$\pi_{\min}=0$, increasing the sigmoid slope $\alpha$ from 2 to 8 shortened the
prefix and increased local error in all 24 model--corpus--$G$ settings. At slope 8,
raising the mass target to 99\% reduced error in every setting, while the median
fraction of channels absent from every selected state fell from 24.9\% to 16.0\%.
We use $\alpha=8$, $\pi_{\min}=0$, and $\tau=0.95$ in the downstream experiments. Table~\ref{tab:prob-scale-sensitivity} in
Appendix~\ref{sec:sensitivity} summarizes the results for all seven probability settings.

At 95\% mass, contiguous groups had lower error than the mean of ten
width-preserving random partitions in 14 of 24 settings. On Llama-3.2-3B at
$G=16$, random grouping reduced mean WikiText-2 error from 0.564 to 0.548.
The percentage of channels absent from every selected state also fell, from
32.6\% to 30.5\%. Appendix~\ref{sec:sensitivity} gives per-model
results and variability across trials.

\subsection{Downstream perplexity at matched projection cost}
\begin{figure}[!htbp]
\centering
\includegraphics[width=\linewidth]{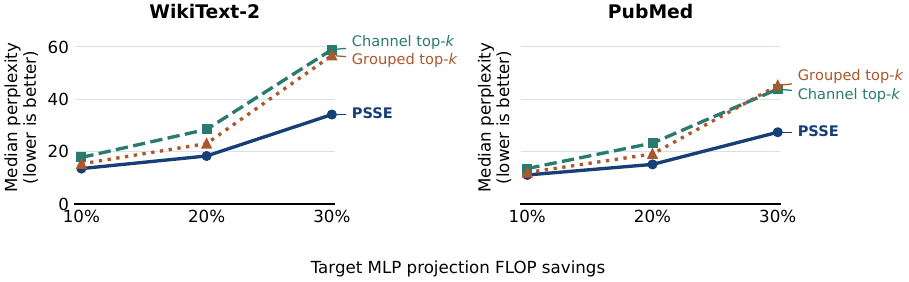}
\caption{Median held-out PPL across six models. At 30\% savings, PSSE, channel top-$k$,
and grouped top-$k$ yield 34.1, 59.0, and 56.8 on WikiText-2 and 27.4, 43.9, and 45.3
on PubMed. Lower is better.}
\label{fig:downstream-ppl}
\end{figure}

\paragraph{Comparison with top-$k$ pruning baselines.}
Both baselines target the same number of channels per layer; PSSE sets each layer's
width from its prefix. At matched projection FLOPs, PSSE yields lower held-out perplexity
(PPL) than both baselines in all 36 settings, with a median reduction of 15.6\% relative
to the better baseline (Table~\ref{tab:main-ppl}, Figure~\ref{fig:downstream-ppl}).

\paragraph{Layer-width allocation ablation.}
We replace the distinct channel count in each layer's 95\%-mass prefix with the
expected active-channel count to set its width target. We hold the scorer, group count,
projection budget, and validation procedure fixed. Prefix-based widths yield lower PPL
in 31 of 36 settings: 17 of 18 on WikiText-2 and 14 of 18 on PubMed. Median reductions
relative to expected-active allocation are 11.2\% and 11.7\%, respectively
(Appendix Table~\ref{tab:width-allocation}).

\begin{table}[!htbp]
\centering
\small
\setlength{\tabcolsep}{5pt}
\caption{Held-out PPL at matched projection FLOPs and validation-selected $G$.
Best top-$k$ is the better of channel and grouped top-$k$. Bold marks the lower PPL.}
\label{tab:main-ppl}
\begin{tabular}{@{}lr l rrr@{}}
\toprule
& & & \multicolumn{3}{c}{PSSE / Best top-$k$ (channel or grouped)} \\
\cmidrule(l){4-6}
Model & $G$ & Dataset & \multicolumn{1}{c}{10\%} & \multicolumn{1}{c}{20\%} & \multicolumn{1}{c}{30\%} \\
\midrule
Llama-3.2-1B & 12 & WikiText-2 & \textbf{15.12} / 16.09 & \textbf{24.23} / 30.93 & \textbf{47.55} / 97.16 \\
 &  & PubMed & \textbf{12.31} / 13.25 & \textbf{19.73} / 29.85 & \textbf{50.85} / 98.29 \\[2pt]
Llama-3.2-3B & 28 & WikiText-2 & \textbf{11.93} / 12.08 & \textbf{17.87} / 22.77 & \textbf{32.69} / 68.01 \\
 &  & PubMed & \textbf{9.39} / 9.59 & \textbf{13.79} / 17.28 & \textbf{28.34} / 44.66 \\[2pt]
Gemma-3-1B & 16 & WikiText-2 & \textbf{22.18} / 27.49 & \textbf{33.23} / 38.28 & \textbf{50.89} / 60.92 \\
 &  & PubMed & \textbf{15.54} / 17.67 & \textbf{23.55} / 25.39 & \textbf{40.42} / 49.28 \\[2pt]
Gemma-3-4B & 12 & WikiText-2 & \textbf{13.89} / 15.12 & \textbf{17.15} / 23.32 & \textbf{22.57} / 40.12 \\
 &  & PubMed & \textbf{12.20} / 13.46 & \textbf{16.37} / 20.88 & \textbf{23.92} / 43.15 \\[2pt]
Qwen3-1.7B & 24 & WikiText-2 & \textbf{13.07} / 14.48 & \textbf{18.73} / 21.37 & \textbf{35.60} / 45.49 \\
 &  & PubMed & \textbf{9.88} / 10.70 & \textbf{13.74} / 16.10 & \textbf{26.42} / 35.50 \\[2pt]
Qwen3-8B & 32 & WikiText-2 & \textbf{9.25} / 9.64 & \textbf{11.71} / 12.23 & \textbf{16.95} / 23.34 \\
 &  & PubMed & \textbf{7.29} / 7.48 & \textbf{9.00} / 9.42 & \textbf{13.12} / 14.31 \\
\bottomrule
\end{tabular}
\end{table}

\subsection{Decoder MLP latency}
\label{sec:cost-accounting}
\begin{table}[!htbp]
\centering
\small
\caption{Decoder MLP latency (ms) and speedup over dense MLPs. We time all methods
on the same GPU model using one 512-token sequence per pass. For each model and
dataset, we sum the MLP times across layers and take the median over 40 passes.
At each savings target, latency is the mean of these 12 medians (six models, two
datasets). Speedup in parentheses is the mean of the 12 within-pair ratios of
dense to pruned median latency.}
\label{tab:mlp-latency}
\begin{tabular}{@{}rrrrr@{}}
\toprule
Target FLOP savings & Dense & PSSE & Channel top-$k$ & Grouped top-$k$ \\
\midrule
10\% & 12.45 & 11.21 ($1.110\times$) & 11.31 ($1.108\times$) & 11.31 ($1.108\times$) \\
20\% & 12.50 & 10.06 ($1.233\times$) & 10.07 ($1.230\times$) & 10.08 ($1.230\times$) \\
30\% & 12.50 & 9.13 ($1.363\times$) & 9.31 ($1.350\times$) & 9.32 ($1.350\times$) \\
\bottomrule
\end{tabular}
\end{table}

\FloatBarrier
Across 1,440 instrumented forward passes with the dense models on the same GPU model,
decoder MLPs accounted for a median of 50.1\% of forward-pass time. Each pass
processed one 512-token sequence. The median share varied by model, from
21.8\% for Gemma-3-1B to 59.5\% for Qwen3-8B.

PSSE reduces summed decoder MLP latency relative to dense MLPs in all 36
model--dataset--budget settings (Appendix Table~\ref{tab:mlp-latency-full}).
The mean paired speedups are $1.11\times$, $1.23\times$, and $1.36\times$
at 10\%, 20\%, and 30\% target MLP projection FLOP savings, respectively.
At 30\% target savings, mean MLP latency falls from 12.50\,ms to 9.13\,ms.
Channel top-$k$ and grouped top-$k$ achieve nearly identical speedups at each
budget (Table~\ref{tab:mlp-latency}). They select different channels but retain
the same number in each layer, so the GPU multiplies matrices of the same
dimensions. PSSE achieves comparable MLP speedups and lower held-out perplexity
than both baselines in every matched comparison (Table~\ref{tab:main-ppl}).

\section{Discussion}
\label{sec:discussion}

\paragraph{Why order joint states rather than rank channels?}
At a given width, a channel ranking identifies a single subset to keep. PSSE instead treats each
subset of channel groups as a state and combines the outputs of several states.
Classical reliability algorithms visit the most probable configurations of working
and failed components first~\citep{li1984performance,lam1986improved}. In an MLP,
the two states of a channel group correspond to its inclusion or exclusion.
Applying the same enumeration order to these configurations yields a
probability-weighted estimate of the MLP output.
At any fixed state count, this ordering captures the most probability mass by
construction. Whether those states also give a better estimate of the dense MLP
output depends on what the selected groups contribute. Figure~\ref{fig:local-ordering}
shows that ordering lowers both median and 99th-percentile output error relative
to uniform sampling at the same state count, with or without bias correction.

\paragraph{Probability mass as a stopping rule.}
With importance correction, the expectation over all states equals the dense
output for any choice of positive group inclusion probabilities. These probabilities
determine the prefix length at a fixed mass target. Layers with more concentrated
distributions reach the target with fewer states. The omitted mass bounds the local
error when multiplied by a bound on the masked-output norm
(Theorem~\ref{thm:mass-truncation}). With group probabilities held fixed, raising
the target from 95\% to 99\% lowered mean error in all 24 model--corpus--$G$
settings (Appendix~\ref{sec:sensitivity}). The scorer determines how costly that
target is to reach. \citet{stahlberg2025outgoing} used the entropy of outgoing
weights to identify neurons to prune. We use that same signal to assign group
inclusion probabilities before seeing a token. In our scorer comparison, ranking
groups by their contribution norms for the current token gave lower local error
but required more states to reach 95\% mass in every model--corpus--$G$ setting
tested (Table~\ref{tab:scorer-comparison}). The weight-based ordering thus gives
a shorter prefix that can also be reused across tokens.

\paragraph{From state prefixes to smaller decoder MLPs.}
The same prefix serves a different purpose in pruning. Rather than evaluate
its weighted sum at inference, we count the distinct channels that appear in it.
Layers whose prefixes span more channels receive a larger share of the projection
budget. Validation determines whether each layer keeps individually ranked channels
or whole groups at its assigned width. We prune once per budget; every token then
uses the smaller dense matrices without scoring channels or enumerating states.
This differs from contextual sparsity in Deja Vu~\citep{liu2023dejavu}, which
predicts which heads and MLP neurons to use from the current input. In our pruned
model the widths vary across layers, while the retained channels within each
layer stay fixed across inputs.
All three pruning methods give comparable decoder MLP speedups. At the same
projection cost, PSSE yields lower held-out perplexity than either top-$k$
baseline in all 36 settings
(Tables~\ref{tab:main-ppl} and~\ref{tab:mlp-latency}).

\paragraph{Broader research directions.}
In machine translation,
\citet{voita2019analyzing} found that specialized heads tended to survive pruning.
Treating subsets of heads as states could identify heads that recur together
in a high-mass prefix. The same analysis could examine which connections recur
across valid sparse subgraphs. In decoding, nucleus sampling draws the next token
from a set chosen by cumulative probability mass~\citep{holtzman2020curious}.
We propose applying the mass criterion to multi-token continuations rather than
individual tokens, and comparing the resulting search with fixed-width beam search
at matched cost. For deployment, we plan to benchmark models with pruned attention
heads and MLP channels on edge devices, measuring peak memory and per-token latency
alongside held-out perplexity.

\section{Conclusion}
\label{sec:conclusion}

Our primary objective was to characterize and control approximation error under a
fixed mask budget. PSSE identifies exactly what the retained masks preserve in a
Transformer MLP. Under independent group inclusions, NORDER-II reaches a prescribed
probability mass with the fewest masks. The exact residual explains why the choice
matters for a given input: each group's contribution to the error depends on how much
of its inclusion probability the selected masks omit. Frozen weights determine the
ordering before any token is evaluated, and the contributions of the retained masks
combine into one reusable coefficient per group.
Across six models and two corpora, the 95\%-mass approximation achieved lower mean
relative error than both uniform estimators in all 1,008 equal-count comparisons.
Even at 32 groups, the mean prefix length ranged from 1,020 to 4,725 masks across
models. At that group count, pooled median relative error was 0.424 for PSSE and
0.652 for a single top-$k$ mask matched to the typical number of active groups in one
state (Table~\ref{tab:main-comparison}). The local comparisons thus quantify what
probability ordering preserves at a given state count.

Our secondary objective was to improve inference efficiency through pruning. We
enumerate masks offline. The distinct channels covered in each layer guide its share
of the pruning budget, and each token then uses the same smaller projection matrices.
At matched MLP projection FLOPs, PSSE achieved lower held-out perplexity than both
top-$k$ baselines in all 36 settings, with a median reduction of 15.6\% relative to
the better baseline (Table~\ref{tab:main-ppl}). Using prefix counts in place of
expected active-channel counts improved perplexity in 31 of 36 settings
(Table~\ref{tab:width-allocation}). All three methods achieved comparable decoder MLP
speedups. At 30\% target projection savings, PSSE's pruned MLPs ran $1.36\times$ as
fast as dense MLPs on average (Table~\ref{tab:mlp-latency}). The practical gain is
better predictive quality at the same projection cost.

These results motivate two next experiments. Approximating successive layers will let
us trace how local residuals accumulate and test whether they predict changes in
held-out perplexity. Timing prompt processing and cached generation separately on
edge devices will establish how much of the measured MLP speedup carries through to
end-to-end inference.

\label{sec:main-end}
\FloatBarrier

\section*{AI Use Statement}

AI assistants supported literature search, code implementation, interpretation of
results, drafting, revision, and language improvement. The authors are responsible for the content
of the submitted manuscript.

\section*{Ethics Statement}

Pruning decoder MLPs reduces the projection cost of inference with pretrained
language models. Lower costs could make these models easier to study for researchers
with limited computing resources. The perplexity increase after pruning varies
across models and datasets, so deployment calls for evaluation on the intended task.

The method inherits the behavioral risks of the underlying language models being
analyzed. It may improve interpretability and diagnostics by exposing which joint
states dominate local computations, but it should not be interpreted as a safety
guarantee or as a mechanism for removing harmful model behavior.

\section*{Reproducibility Statement}

Sections~\ref{sec:method} and~\ref{sec:exp} describe the state construction,
channel selection, and matched-FLOP comparisons.
Appendices~\ref{sec:sensitivity} and~\ref{sec:packed-protocol}
give the sensitivity settings and the calibration, scoring, and timing protocols.
Appendix~\ref{sec:controlled-illustrations} reports controlled checks on the state
ordering, and Appendix~\ref{sec:bound} proves Theorem~\ref{thm:mass-truncation}.

\section*{Acknowledgment}

This work was performed as part of the authors' work at Elohax Ltd.

\section*{Conflict of Interest}

A patent application related to this work has been filed with the United States Patent and Trademark Office.

\bibliography{references}
\bibliographystyle{template/iclr2027_conference}

\clearpage
\appendix
\section*{Appendix: Supplementary Material}

\section{Worked Example: From Three MLP Groups to Joint States}
\label{sec:ffn-example}

Partition the MLP hidden coordinates into three disjoint groups $I_1,I_2,I_3$, and let
$c_g(x)$ denote the output contribution of the coordinates in $I_g$. The dense output is
$y(x)=c_1(x)+c_2(x)+c_3(x)$. Choose arbitrary inclusion probabilities
$\pi_g\in(0,1]$ and independent $z_g\sim\mathrm{Bernoulli}(\pi_g)$. Written as
$z_1z_2z_3$, the eight auxiliary masks are
\[
\{000,001,010,011,100,101,110,111\}.
\]
For example, $z=(1,0,1)$ produces
\begin{equation}
F(x,(1,0,1))=\frac{c_1(x)}{\pi_1}+\frac{c_3(x)}{\pi_3}.
\end{equation}
Summing over all eight states gives
\begin{equation}
\mathbb{E}_z[F(x,z)]
=\sum_{g=1}^3\frac{\mathbb{E}[z_g]}{\pi_g}c_g(x)
=y(x).
\end{equation}
The full auxiliary expectation is therefore exact; approximation begins only when it
is truncated to a probability-ordered prefix. Appendix~\ref{sec:controlled-illustrations}
instantiates the same construction in a trained one-hidden-layer network.

\section{Approximation Guarantee}
\label{sec:bound}

{
\begin{proof}[Proof of Theorem~\ref{thm:mass-truncation}]
Let $\rho$ be the retained probability mass. The residual is the probability-weighted
sum of the omitted masked outputs. Each has norm at most $B(x)$, so the triangle
inequality bounds the residual norm by $(1-\rho)B(x)$, even if the state budget is
exhausted before the target is reached. When $\rho\ge\tau$, this is at most $(1-\tau)B(x)$.
\end{proof}

\begin{proposition}[Exact truncation error]
\label{prop:exact-truncation-error}
Fix an input $x$ and any retained set of states $\mathcal{P}$ with mass $\rho$.
The coefficients $a_g=a_g(\mathcal{P},x)$ in Eq.~\eqref{eq:coefficient-collapsed-prefix}
satisfy $0\le a_g\le 1$. The residual is
$y(x)-\hat y_{\mathcal{P}}(x)=\sum_g(1-a_g)c_g(x)$, and its norm satisfies
\[
\|y(x)-\hat y_{\mathcal{P}}(x)\|
\le \sum_g(1-a_g)\|c_g(x)\|
\le (1-\rho)\sum_g\frac{\|c_g(x)\|}{\pi_g}.
\]
\end{proposition}
\begin{proof}
The retained and omitted states in which $z_g=1$ have total mass $\pi_g>0$.
The retained part is $\pi_g a_g$, so $0\le a_g\le 1$. The omitted part is at
most $1-\rho$, giving $1-a_g\le (1-\rho)/\pi_g$. Subtracting
Eq.~\eqref{eq:coefficient-collapsed-prefix} from $y(x)=\sum_g c_g(x)$ gives
the residual. The triangle inequality and the bound on $1-a_g$ give the two inequalities.

Let $U=\{g:a_g>0\}$. Retaining every group in $U$ at full weight leaves the
residual $\sum_{g\notin U}c_g(x)$. Its norm bound $\sum_{g\notin U}\|c_g(x)\|$
is no larger than $\sum_g(1-a_g)\|c_g(x)\|$, since $a_g=0$ outside $U$.
The actual errors need not follow the same ordering because group vectors can cancel.
\end{proof}

The base state includes all groups with $\pi_g\ge 1/2$. For a group with
$\pi_g<1/2$, a most probable state that includes it flips only that group from
the base state. Its probability is $P(\mathrm{base}\mid x)\,\pi_g/(1-\pi_g)$,
which increases with $\pi_g$. A probability-ordered prefix therefore covers the
highest-probability groups up to ties.
}

\section{Controlled Joint-State Illustrations}
\label{sec:controlled-illustrations}

This appendix gives two controlled illustrations supporting the methodology in
Section~\ref{sec:method} and Appendix~\ref{sec:ffn-example}. The first verifies that NORDER-II
selects the same ordered high-probability states as exhaustive enumeration in
independent binary state spaces. The second applies the same construction to
groups of hidden units in a trained one-hidden-layer classifier.

\paragraph{Correctness against exhaustive enumeration.}
We sample independent factor probabilities and compare NORDER-II with exhaustive
enumeration for $G \in \{4,8,12,16\}$ and
$\tau \in \{0.80,0.90,0.95,0.99\}$. Exhaustive enumeration first lists all $2^G$
states, sorts them by probability, and selects the smallest ordered set whose
cumulative mass reaches $\tau$. In all configurations, NORDER-II selects the same
ordered set without sorting the full state space, preserves non-increasing probability
order, produces no duplicate states, and stops as soon as the cumulative mass reaches
$\tau$. Table~\ref{tab:controlled-correctness} reports representative results at
$\tau=0.95$. The table gives the prefix length at the mass target and its fraction
of the full state space. Exact agreement with exhaustive sorting checks both the
order of the retained states and the stopping point.

\begin{table}[ht]
\centering
\footnotesize
\caption{Controlled exhaustive-enumeration check at $\tau=0.95$. NORDER-II selects the same ordered states as exhaustive sorting in all rows.}
\label{tab:controlled-correctness}
\begin{tabular}{rrrrc}
\toprule
$G$ & $2^G$ & Selected & Frac. & Match \\
\midrule
4  & 16    & 10    & 0.625 & Yes \\
8  & 256   & 113   & 0.441 & Yes \\
12 & 4096  & 1774  & 0.433 & Yes \\
16 & 65536 & 12684 & 0.194 & Yes \\
\bottomrule
\end{tabular}
\end{table}

\begin{figure}[ht]
\centering
\includegraphics[width=0.85\linewidth]{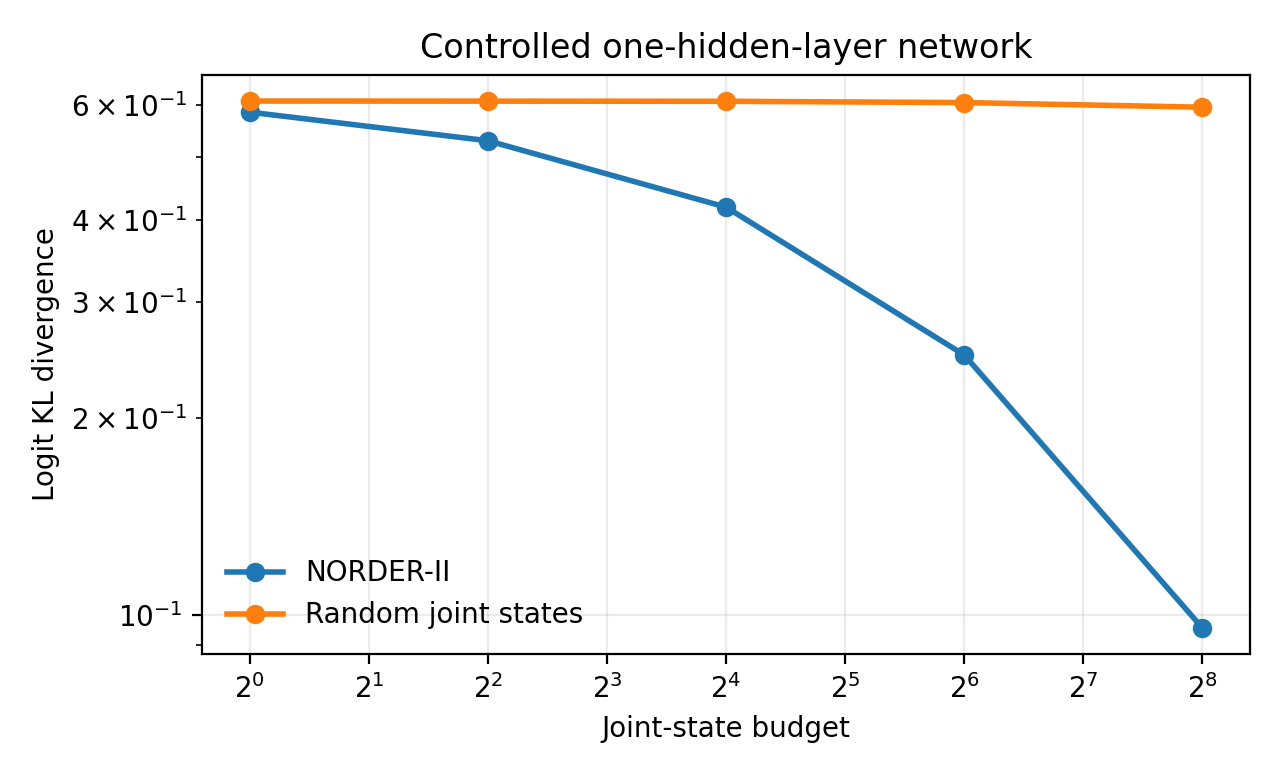}
\caption{Mean KL divergence from the dense classifier's output distribution
to its approximation, by number of states. Probability-ordered prefixes yield lower
divergence than uniform samples of the same size.}
\label{fig:controlled-trained-mlp}
\end{figure}

\paragraph{Trained one-hidden-layer network.}
We train a one-hidden-layer classifier on a two-class nonlinear classification problem
and then freeze its parameters. Hidden units are partitioned into groups, inducing a
$2^G$ joint binary state space over group inclusion patterns. The dense validation
accuracy is $0.965$. We apply the construction in Appendix~\ref{sec:ffn-example}
to these learned representations and compare the resulting class probabilities.
Figure~\ref{fig:controlled-trained-mlp} reports
$D_{\mathrm{KL}}(p_{\mathrm{dense}}\Vert p_{\mathrm{approx}})$, where each
distribution is the softmax of the corresponding logits. Probability-ordered
prefixes give lower mean divergence than uniform samples at every state count tested.

\FloatBarrier
{
\section{Local Approximation and Sensitivity to the Group Scorer}
\label{sec:scorer-comparison}}

Table~\ref{tab:main-comparison} shows how local error varies by model and corpus
(panel A) and across individual prompt--layer comparisons (panel B).
To compute the mean number of active groups per state, we count the groups included
in each selected mask and give every mask in every layer's prefix equal weight.
The 168 layer prefixes contain 821, 5,635, and 355,675 states in total at
$G=8$, 16, and 32, respectively. Each layer's prefix is shared across prompts and corpora.
For the \emph{Single top-$k$ mask (mean active count)} baseline, we round the mean active
count in each layer's prefix to the nearest integer $k$, rounding half up. The mask retains
the $k$ groups with the highest inclusion probabilities (ties favor lower group indices)
and sums their original contributions $c_g(x)$.

\begin{table}[H]
\centering
\small
\setlength{\tabcolsep}{2pt}
\caption{Relative error of local MLP output approximations (lower is better). Uniform and HT
use the same random draw at PSSE's state count, without and with Horvitz--Thompson correction, respectively.
The single top-$k$ mask retains each layer's rounded mean number of active groups per selected state.
Panel A averages layer means within each model. The column $m$ gives the mean prefix length at 95\% mass.
Panel B pools all 86,016 prompt--layer errors per method and $G$ (256 prompts per layer,
168 layers, two corpora) before computing the median and 99th percentile (P99).}
\label{tab:main-comparison}
\begin{tabular*}{\linewidth}{@{\extracolsep{\fill}}lrrrrrrrrrr@{}}
\toprule
\multicolumn{11}{@{}l}{\textbf{A. Mean error by model and corpus}} \\
& & & \multicolumn{4}{c}{WikiText-2} & \multicolumn{4}{c}{PubMed} \\
\cmidrule(lr){4-7}\cmidrule(l){8-11}
Model & $G$ & $m$ & PSSE & Uniform & HT & Top-$k$ mask & PSSE & Uniform & HT & Top-$k$ mask \\
\midrule
Llama-3.2-1B & 8 & 4.7 & 0.598 & 0.985 & 2.382 & 0.695 & 0.605 & 0.985 & 2.379 & 0.704 \\
 & 16 & 27.1 & 0.549 & 1.000 & 2.847 & 0.712 & 0.553 & 1.000 & 2.797 & 0.715 \\
 & 32 & 2,731.9 & 0.474 & 1.000 & 2.276 & 0.704 & 0.481 & 1.000 & 2.362 & 0.705 \\
Llama-3.2-3B & 8 & 3.7 & 0.600 & 0.989 & 2.422 & 0.680 & 0.597 & 0.989 & 2.431 & 0.679 \\
 & 16 & 19.6 & 0.564 & 1.000 & 3.815 & 0.702 & 0.564 & 1.000 & 3.933 & 0.705 \\
 & 32 & 1,667.5 & 0.474 & 1.000 & 18.104 & 0.704 & 0.477 & 1.000 & 21.065 & 0.710 \\
Gemma-3-1B & 8 & 4.8 & 0.429 & 0.982 & 2.049 & 0.538 & 0.443 & 0.983 & 2.085 & 0.552 \\
 & 16 & 19.2 & 0.387 & 1.000 & 2.548 & 0.511 & 0.403 & 1.000 & 2.512 & 0.528 \\
 & 32 & 1,019.8 & 0.329 & 1.000 & 1.482 & 0.507 & 0.346 & 1.000 & 1.575 & 0.530 \\
Gemma-3-4B & 8 & 5.9 & 0.392 & 0.978 & 1.922 & 0.517 & 0.405 & 0.978 & 1.963 & 0.534 \\
 & 16 & 30.6 & 0.376 & 1.000 & 2.107 & 0.501 & 0.387 & 1.000 & 2.249 & 0.515 \\
 & 32 & 1,259.0 & 0.327 & 1.000 & 1.727 & 0.488 & 0.343 & 1.000 & 1.596 & 0.505 \\
Qwen3-1.7B & 8 & 5.3 & 0.546 & 0.981 & 2.349 & 0.679 & 0.548 & 0.981 & 2.355 & 0.679 \\
 & 16 & 41.6 & 0.513 & 0.999 & 2.764 & 0.677 & 0.520 & 0.999 & 2.751 & 0.685 \\
 & 32 & 4,725.0 & 0.458 & 1.000 & 1.402 & 0.668 & 0.464 & 1.000 & 1.434 & 0.676 \\
Qwen3-8B & 8 & 4.7 & 0.562 & 0.984 & 2.390 & 0.665 & 0.566 & 0.985 & 2.382 & 0.668 \\
 & 16 & 54.1 & 0.510 & 0.999 & 2.582 & 0.680 & 0.514 & 0.999 & 2.646 & 0.681 \\
 & 32 & 1,768.2 & 0.460 & 1.000 & 2.786 & 0.667 & 0.459 & 1.000 & 3.165 & 0.668 \\
\end{tabular*}\par\medskip
\begin{tabular*}{\linewidth}{@{\extracolsep{\fill}}lrrrrrrrr@{}}
\midrule
\multicolumn{9}{@{}l}{\textbf{B. Pooled prompt--layer error percentiles}} \\
& \multicolumn{2}{c}{PSSE} & \multicolumn{2}{c}{Uniform} & \multicolumn{2}{c}{HT} & \multicolumn{2}{c}{Single top-$k$ mask} \\
\cmidrule(lr){2-3}\cmidrule(lr){4-5}\cmidrule(lr){6-7}\cmidrule(l){8-9}
$G$ & Median & P99 & Median & P99 & Median & P99 & Median & P99 \\
\midrule
8 & 0.523 & 0.815 & 1.000 & 1.000 & 1.000 & 25.795 & 0.648 & 0.862 \\
16 & 0.487 & 0.766 & 1.000 & 1.000 & 1.000 & 28.839 & 0.659 & 0.874 \\
32 & 0.424 & 0.708 & 1.000 & 1.000 & 1.000 & 4.529 & 0.652 & 0.867 \\
\bottomrule
\end{tabular*}
\end{table}

How does the choice of group score affect the number of states needed to reach
95\% mass and the error of the resulting MLP estimate? As in Table~\ref{tab:main-comparison},
we evaluate both scorers at the last non-padding token of each of 256 prompts per
dataset, across all 168 MLP layers in the six models. The weight-entropy
score is fixed within a layer. The contribution-norm score $\|c_g(x)\|_2$ varies by
token and requires computing all $G$ group outputs. Both use contiguous groups and
the same standardized probability transform ($\alpha=8$, $\beta=0$, $\pi_{\min}=0$).
For each scorer we take its shortest prefix reaching 95\% of its own probability
mass, with no state cap.

The contribution-norm scorer has lower mean local error and requires more states
in all 24 model, corpus and group-count combinations. For example, on Gemma-3-4B
with $G=16$ on WikiText-2, mean error falls from 0.376 to 0.251 while mean prefix
length rises from 30.6 to 159.5 states. Weight entropy lets us compute the shorter
prefix once per layer and reuse it across tokens.

\begin{table}[H]

\centering
\small
\setlength{\tabcolsep}{4pt}
\caption{Local scorer comparison at 95\% mass. $m$ is the mean prefix length in
states. Error is the mean relative $\ell_2$ error against the dense MLP output.
Both quantities average over prompts and layers within a model.
Weight denotes weight entropy and Norm denotes contribution norm.}
\label{tab:scorer-comparison}
\begin{tabular}{@{}lrrrrr@{\hspace{12pt}}rrrr@{}}
\toprule
& & \multicolumn{4}{c}{WikiText-2} & \multicolumn{4}{c}{PubMed} \\
\cmidrule(lr){3-6}\cmidrule(l){7-10}
& & \multicolumn{2}{c}{$m$} & \multicolumn{2}{c}{Error} &
\multicolumn{2}{c}{$m$} & \multicolumn{2}{c}{Error} \\
Model & $G$ & Weight & Norm & Weight & Norm & Weight & Norm & Weight & Norm \\
\midrule
Llama-3.2-1B & 16 & 27.1 & 70.9 & 0.549 & 0.445 & 27.1 & 59.8 & 0.553 & 0.456 \\
 & 32 & 2,731.9 & $1.56\times 10^{5}$ & 0.474 & 0.306 & 2,731.9 & $3.52\times 10^{4}$ & 0.481 & 0.313 \\
Llama-3.2-3B & 16 & 19.6 & 64.8 & 0.564 & 0.444 & 19.6 & 66.5 & 0.564 & 0.447 \\
 & 32 & 1,667.5 & $8.33\times 10^{4}$ & 0.474 & 0.303 & 1,667.5 & $7.36\times 10^{4}$ & 0.477 & 0.307 \\
Gemma-3-1B & 16 & 19.2 & 86.1 & 0.387 & 0.282 & 19.2 & 86.5 & 0.403 & 0.299 \\
 & 32 & 1,019.8 & $1.17\times 10^{5}$ & 0.329 & 0.172 & 1,019.8 & $1.28\times 10^{5}$ & 0.346 & 0.186 \\
Gemma-3-4B & 16 & 30.6 & 159.5 & 0.376 & 0.251 & 30.6 & 156.8 & 0.387 & 0.261 \\
 & 32 & 1,259.0 & $7.09\times 10^{5}$ & 0.327 & 0.153 & 1,259.0 & $5.86\times 10^{5}$ & 0.343 & 0.160 \\
Qwen3-1.7B & 16 & 41.6 & 57.8 & 0.513 & 0.387 & 41.6 & 68.1 & 0.520 & 0.374 \\
 & 32 & 4,725.0 & $7.72\times 10^{4}$ & 0.458 & 0.262 & 4,725.0 & $1.22\times 10^{5}$ & 0.464 & 0.248 \\
Qwen3-8B & 16 & 54.1 & 68.8 & 0.510 & 0.402 & 54.1 & 76.4 & 0.514 & 0.392 \\
 & 32 & 1,768.2 & $1.57\times 10^{5}$ & 0.460 & 0.279 & 1,768.2 & $2.01\times 10^{5}$ & 0.459 & 0.263 \\
\bottomrule
\end{tabular}
\end{table}

\section{Sensitivity Protocols and Results}
\label{sec:sensitivity}
We hold the model, input token, frozen channel scores, and dense reference
fixed within each paired comparison. We randomly reassign channels within each
layer while preserving each group's size.
We use the same ten random partitions for each layer and $G$ on both corpora.
We recompute the group scores for each partition, average prompt errors within
layers, and then average layers within each trial. Table~\ref{tab:partition} gives
the mean and sample standard deviation across trials.

\begin{table}[htbp]
\centering
\footnotesize
\setlength{\tabcolsep}{3pt}
\caption{Local error for contiguous groups and ten random partitions (mean $\pm$ sample SD across trials). The last two columns give the mean percentage of channels absent from every state in the prefix. For a given partition, the weight-based prefix is the same on both corpora.}
\label{tab:partition}
\begin{tabular}{@{}lrrrrrrr@{}}
\toprule
& & \multicolumn{2}{c}{WikiText-2 error} & \multicolumn{2}{c}{PubMed error} & \multicolumn{2}{c}{Channels omitted (\%)} \\
\cmidrule(lr){3-4}\cmidrule(lr){5-6}\cmidrule(l){7-8}
Model & $G$ & Contiguous & Random & Contiguous & Random & Contiguous & Random \\
\midrule
Llama-3.2-1B & 16 & 0.5495 & 0.5512 $\pm$ 0.0158 & 0.5526 & 0.5543 $\pm$ 0.0159 & 30.9 & 31.0 \\
 & 32 & 0.4741 & 0.4664 $\pm$ 0.0118 & 0.4815 & 0.4722 $\pm$ 0.0122 & 20.5 & 18.8 \\
Llama-3.2-3B & 16 & 0.5641 & 0.5475 $\pm$ 0.0148 & 0.5645 & 0.5508 $\pm$ 0.0129 & 32.6 & 30.5 \\
 & 32 & 0.4736 & 0.4753 $\pm$ 0.0083 & 0.4772 & 0.4786 $\pm$ 0.0071 & 19.3 & 19.6 \\
Gemma-3-1B & 16 & 0.3868 & 0.3896 $\pm$ 0.0120 & 0.4031 & 0.4100 $\pm$ 0.0131 & 31.5 & 31.4 \\
 & 32 & 0.3290 & 0.3339 $\pm$ 0.0114 & 0.3457 & 0.3537 $\pm$ 0.0120 & 19.0 & 20.9 \\
Gemma-3-4B & 16 & 0.3764 & 0.3731 $\pm$ 0.0095 & 0.3867 & 0.3895 $\pm$ 0.0090 & 31.6 & 30.6 \\
 & 32 & 0.3267 & 0.3207 $\pm$ 0.0061 & 0.3428 & 0.3345 $\pm$ 0.0079 & 20.6 & 19.7 \\
Qwen3-1.7B & 16 & 0.5129 & 0.5163 $\pm$ 0.0133 & 0.5195 & 0.5199 $\pm$ 0.0119 & 28.6 & 29.6 \\
 & 32 & 0.4576 & 0.4537 $\pm$ 0.0125 & 0.4641 & 0.4585 $\pm$ 0.0115 & 20.1 & 19.3 \\
Qwen3-8B & 16 & 0.5104 & 0.5233 $\pm$ 0.0088 & 0.5136 & 0.5235 $\pm$ 0.0102 & 29.0 & 30.7 \\
 & 32 & 0.4596 & 0.4579 $\pm$ 0.0090 & 0.4588 & 0.4593 $\pm$ 0.0085 & 21.2 & 20.1 \\
\bottomrule
\end{tabular}
\end{table}

Table~\ref{tab:prob-scale-sensitivity} summarizes how the probability settings
affect prefix length, the fraction of channels omitted from the prefix, and local error for contiguous groups.
Tables~\ref{tab:sweep-llama}--\ref{tab:sweep-qwen} give the results for each model family.
We vary the sigmoid slope $\alpha$ with $\pi_{\min}=0$, compare probability floors
at $\alpha=2$ and $8$, and vary the mass target $\tau$ at $\alpha=8$.
Every reported prefix is computed exactly and reaches its mass target.
For numerical stability, we clip the Bernoulli
parameters to $[10^{-8},1-10^{-8}]$ before computing log probabilities and
importance weights.
\begin{table}[htbp]
\centering
\small
\setlength{\tabcolsep}{4pt}
\caption{Probability settings for contiguous groups. Each reported outcome is the median of 24 means, one per model, corpus, and group count. A channel is omitted if it appears in none of the selected states. The pruning experiments set a separate budget for MLP projection FLOPs.}
\label{tab:prob-scale-sensitivity}
\begin{tabular}{@{}rrrrrr@{}}
\toprule
Slope $\alpha$ & Floor $\pi_{\min}$ & Mass $\tau$ & States/layer & Channels omitted (\%) & Relative error \\
\midrule
2 & 0.00 & 0.95 & $1.08\times 10^{7}$ & 0.0 & 0.094 \\
4 & 0.00 & 0.95 & $6.19\times 10^{4}$ & 2.8 & 0.222 \\
8 & 0.00 & 0.95 & 536.9 & 24.9 & 0.469 \\
2 & 0.05 & 0.95 & $5.19\times 10^{7}$ & 0.0 & 0.075 \\
8 & 0.05 & 0.95 & $1.68\times 10^{5}$ & 0.0 & 0.100 \\
8 & 0.00 & 0.90 & 256.2 & 30.7 & 0.527 \\
8 & 0.00 & 0.99 & 2,066.9 & 16.0 & 0.353 \\
\bottomrule
\end{tabular}
\end{table}

\begin{table}[htbp]
\centering
\footnotesize
\setlength{\tabcolsep}{2.8pt}
\caption{Sensitivity to the probability settings for Llama-3.2-1B and Llama-3.2-3B. We report mean prefix length ($m$), the percentage of channels absent from every selected state, and mean local relative error on WikiText-2 (WT) and PubMed (PM).}
\label{tab:sweep-llama}
\begin{tabular}{@{}rrrrrrrrrrrr@{}}
\toprule
& & & & \multicolumn{4}{c}{Llama-3.2-1B} & \multicolumn{4}{c}{Llama-3.2-3B} \\
\cmidrule(lr){5-8}\cmidrule(l){9-12}
$G$ & $\alpha$ & Floor & $\tau$ & $m$ & \% omitted & WT & PM & $m$ & \% omitted & WT & PM \\
\midrule
16 & 2 & 0.00 & 0.95 & 4,372.1 & 0.0 & 0.118 & 0.120 & 4,208.2 & 0.0 & 0.124 & 0.125 \\
 & 4 & 0.00 & 0.95 & 309.8 & 5.1 & 0.301 & 0.305 & 258.1 & 6.2 & 0.316 & 0.319 \\
 & 8 & 0.00 & 0.95 & 27.1 & 30.9 & 0.549 & 0.553 & 19.6 & 32.6 & 0.564 & 0.564 \\
 & 2 & 0.05 & 0.95 & 9,835.9 & 0.0 & 0.089 & 0.090 & 9,587.9 & 0.0 & 0.091 & 0.092 \\
 & 8 & 0.05 & 0.95 & 618.8 & 0.0 & 0.130 & 0.130 & 535.4 & 0.0 & 0.129 & 0.129 \\
 & 8 & 0.00 & 0.90 & 16.9 & 36.7 & 0.604 & 0.605 & 12.1 & 37.5 & 0.607 & 0.607 \\
 & 8 & 0.00 & 0.99 & 67.1 & 21.5 & 0.444 & 0.450 & 49.6 & 21.0 & 0.445 & 0.448 \\
32 & 2 & 0.00 & 0.95 & $3.19\times 10^{7}$ & 0.0 & 0.090 & 0.091 & $2.77\times 10^{7}$ & 0.0 & 0.094 & 0.095 \\
 & 4 & 0.00 & 0.95 & $3.70\times 10^{5}$ & 0.0 & 0.209 & 0.212 & $2.71\times 10^{5}$ & 0.4 & 0.212 & 0.214 \\
 & 8 & 0.00 & 0.95 & 2,731.9 & 20.5 & 0.474 & 0.481 & 1,667.5 & 19.3 & 0.474 & 0.477 \\
 & 2 & 0.05 & 0.95 & $1.29\times 10^{8}$ & 0.0 & 0.074 & 0.074 & $1.20\times 10^{8}$ & 0.0 & 0.075 & 0.075 \\
 & 8 & 0.05 & 0.95 & $5.08\times 10^{5}$ & 0.0 & 0.099 & 0.100 & $3.94\times 10^{5}$ & 0.0 & 0.100 & 0.100 \\
 & 8 & 0.00 & 0.90 & 1,261.0 & 24.4 & 0.526 & 0.533 & 771.3 & 24.4 & 0.539 & 0.541 \\
 & 8 & 0.00 & 0.99 & $1.11\times 10^{4}$ & 11.3 & 0.352 & 0.358 & 6,853.2 & 10.6 & 0.352 & 0.357 \\
\bottomrule
\end{tabular}
\end{table}

\begin{table}[htbp]
\centering
\footnotesize
\setlength{\tabcolsep}{2.8pt}
\caption{Sensitivity to the probability settings for Gemma-3-1B and Gemma-3-4B. We report mean prefix length ($m$), the percentage of channels absent from every selected state, and mean local relative error on WikiText-2 (WT) and PubMed (PM).}
\label{tab:sweep-gemma3_1b}
\begin{tabular}{@{}rrrrrrrrrrrr@{}}
\toprule
& & & & \multicolumn{4}{c}{Gemma-3-1B} & \multicolumn{4}{c}{Gemma-3-4B} \\
\cmidrule(lr){5-8}\cmidrule(l){9-12}
$G$ & $\alpha$ & Floor & $\tau$ & $m$ & \% omitted & WT & PM & $m$ & \% omitted & WT & PM \\
\midrule
16 & 2 & 0.00 & 0.95 & 3,976.1 & 0.0 & 0.099 & 0.101 & 4,211.7 & 0.0 & 0.093 & 0.095 \\
 & 4 & 0.00 & 0.95 & 231.5 & 7.0 & 0.239 & 0.248 & 292.1 & 7.7 & 0.226 & 0.235 \\
 & 8 & 0.00 & 0.95 & 19.2 & 31.5 & 0.387 & 0.403 & 30.6 & 31.6 & 0.376 & 0.387 \\
 & 2 & 0.05 & 0.95 & 9,174.8 & 0.0 & 0.076 & 0.077 & 9,661.8 & 0.0 & 0.073 & 0.074 \\
 & 8 & 0.05 & 0.95 & 554.7 & 0.0 & 0.101 & 0.103 & 722.4 & 0.0 & 0.100 & 0.102 \\
 & 8 & 0.00 & 0.90 & 12.2 & 34.6 & 0.430 & 0.446 & 19.6 & 35.1 & 0.423 & 0.432 \\
 & 8 & 0.00 & 0.99 & 47.0 & 23.6 & 0.318 & 0.333 & 72.0 & 23.7 & 0.304 & 0.315 \\
32 & 2 & 0.00 & 0.95 & $2.17\times 10^{7}$ & 0.0 & 0.076 & 0.078 & $2.30\times 10^{7}$ & 0.0 & 0.073 & 0.074 \\
 & 4 & 0.00 & 0.95 & $1.23\times 10^{5}$ & 0.1 & 0.157 & 0.164 & $1.28\times 10^{5}$ & 0.0 & 0.144 & 0.150 \\
 & 8 & 0.00 & 0.95 & 1,019.8 & 19.0 & 0.329 & 0.346 & 1,259.0 & 20.6 & 0.327 & 0.343 \\
 & 2 & 0.05 & 0.95 & $1.04\times 10^{8}$ & 0.0 & 0.064 & 0.065 & $1.08\times 10^{8}$ & 0.0 & 0.062 & 0.063 \\
 & 8 & 0.05 & 0.95 & $3.36\times 10^{5}$ & 0.0 & 0.083 & 0.084 & $4.01\times 10^{5}$ & 0.0 & 0.080 & 0.082 \\
 & 8 & 0.00 & 0.90 & 478.9 & 24.8 & 0.393 & 0.409 & 622.9 & 27.7 & 0.385 & 0.401 \\
 & 8 & 0.00 & 0.99 & 4,007.0 & 10.9 & 0.227 & 0.242 & 4,586.6 & 9.8 & 0.225 & 0.238 \\
\bottomrule
\end{tabular}
\end{table}

\begin{table}[htbp]
\centering
\footnotesize
\setlength{\tabcolsep}{2.8pt}
\caption{Sensitivity to the probability settings for Qwen3-1.7B and Qwen3-8B. We report mean prefix length ($m$), the percentage of channels absent from every selected state, and mean local relative error on WikiText-2 (WT) and PubMed (PM).}
\label{tab:sweep-qwen}
\begin{tabular}{@{}rrrrrrrrrrrr@{}}
\toprule
& & & & \multicolumn{4}{c}{Qwen3-1.7B} & \multicolumn{4}{c}{Qwen3-8B} \\
\cmidrule(lr){5-8}\cmidrule(l){9-12}
$G$ & $\alpha$ & Floor & $\tau$ & $m$ & \% omitted & WT & PM & $m$ & \% omitted & WT & PM \\
\midrule
16 & 2 & 0.00 & 0.95 & 5,062.1 & 0.0 & 0.122 & 0.126 & 4,933.8 & 0.0 & 0.118 & 0.119 \\
 & 4 & 0.00 & 0.95 & 473.6 & 5.8 & 0.289 & 0.294 & 463.7 & 6.4 & 0.312 & 0.316 \\
 & 8 & 0.00 & 0.95 & 41.6 & 28.6 & 0.513 & 0.520 & 54.1 & 29.0 & 0.510 & 0.514 \\
 & 2 & 0.05 & 0.95 & $1.09\times 10^{4}$ & 0.0 & 0.087 & 0.089 & $1.08\times 10^{4}$ & 0.0 & 0.089 & 0.089 \\
 & 8 & 0.05 & 0.95 & 721.1 & 0.0 & 0.124 & 0.125 & 819.4 & 0.0 & 0.124 & 0.125 \\
 & 8 & 0.00 & 0.90 & 24.4 & 33.7 & 0.564 & 0.568 & 33.4 & 33.7 & 0.558 & 0.562 \\
 & 8 & 0.00 & 0.99 & 107.7 & 19.6 & 0.400 & 0.409 & 126.7 & 20.7 & 0.417 & 0.421 \\
32 & 2 & 0.00 & 0.95 & $4.86\times 10^{7}$ & 0.0 & 0.091 & 0.093 & $2.70\times 10^{7}$ & 0.0 & 0.093 & 0.092 \\
 & 4 & 0.00 & 0.95 & $1.47\times 10^{6}$ & 0.1 & 0.214 & 0.218 & $2.19\times 10^{5}$ & 0.3 & 0.216 & 0.215 \\
 & 8 & 0.00 & 0.95 & 4,725.0 & 20.1 & 0.458 & 0.464 & 1,768.2 & 21.2 & 0.460 & 0.459 \\
 & 2 & 0.05 & 0.95 & $1.42\times 10^{8}$ & 0.0 & 0.074 & 0.074 & $1.16\times 10^{8}$ & 0.0 & 0.074 & 0.074 \\
 & 8 & 0.05 & 0.95 & $4.69\times 10^{5}$ & 0.0 & 0.097 & 0.097 & $4.36\times 10^{5}$ & 0.0 & 0.097 & 0.096 \\
 & 8 & 0.00 & 0.90 & 1,897.0 & 26.1 & 0.523 & 0.527 & 837.0 & 25.1 & 0.514 & 0.513 \\
 & 8 & 0.00 & 0.99 & $2.48\times 10^{4}$ & 11.8 & 0.341 & 0.349 & 6,916.2 & 12.3 & 0.354 & 0.356 \\
\bottomrule
\end{tabular}
\end{table}

\FloatBarrier
\section{Evaluation Protocol and Results for Pruned MLPs}
\label{sec:packed-protocol}
\begin{figure}[!htbp]
\centering
\includegraphics[width=\linewidth]{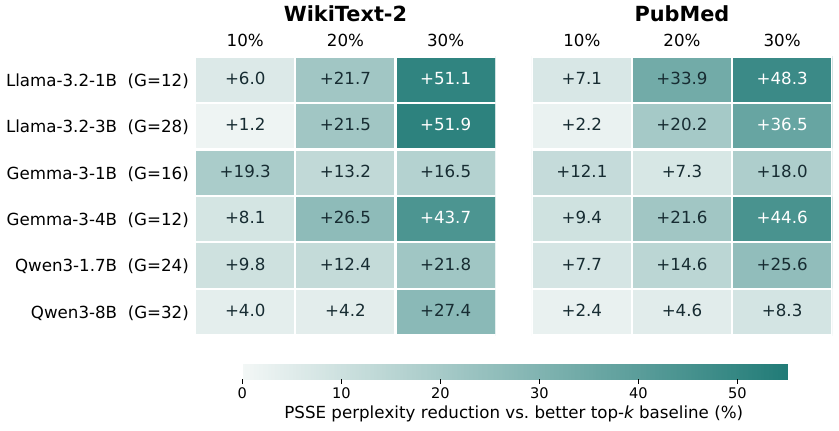}
\caption{Held-out perplexity reduction for each model, dataset, and budget.
Entries give $100(1-\mathrm{PPL}_{\rm PSSE}/\mathrm{PPL}_{\rm best})$, where
$\mathrm{PPL}_{\rm best}$ is the lower of the two top-$k$ perplexities.
Positive values favor PSSE. Each row uses the same selected $G$ for all methods.}
\label{fig:downstream-heatmap}
\end{figure}

For each model and $G$, we determine channel versus group selection
using 4,096 next-token prediction targets from WikiText-2 validation data. At each
savings target, we allocate widths from prefixes that reach 95\% mass.
Keeping the other layers dense, we measure each rule's mean next-token negative
log-likelihood at the layer's three assigned widths. We select the rule with the
lower average loss across those widths, giving each width equal weight. Ties favor
whole groups. With all layers pruned, we score the candidate group counts on
a separate block of 4,096 WikiText-2 validation targets. We average PSSE's next-token loss
equally across budgets and choose the smallest $G$ attaining the minimum.
All three methods use this group count on both test corpora.
All three methods keep the original weights of the retained channels in the pruned MLPs.
Figure~\ref{fig:downstream-heatmap} shows the paired PPL reductions and
Table~\ref{tab:downstream} gives the underlying perplexities.

\begin{table}[!htbp]
\centering
\footnotesize
\setlength{\tabcolsep}{3pt}
\caption{Held-out perplexity of models with pruned MLPs at matched projection FLOPs using one validation-selected $G$ per model. Dense gives the original model's PPL. Save\,\% is the target reduction in MLP projection FLOPs. Ch. and Grp. denote channel top-$k$ and grouped top-$k$. Bold marks the lowest PPL among the three pruning methods on each corpus.}
\label{tab:downstream}
\begin{tabular}{@{}lrrrrrrrrrr@{}}
\toprule
& & & \multicolumn{4}{c}{WikiText-2} & \multicolumn{4}{c}{PubMed} \\
\cmidrule(lr){4-7}\cmidrule(l){8-11}
Model & $G$ & Save\,\% & Dense & PSSE & Ch. & Grp. & Dense & PSSE & Ch. & Grp. \\
\midrule
Llama-3.2-1B & 12 & 10 & 11.10 & \textbf{15.12} & 85.77 & 16.09 & 9.01 & \textbf{12.31} & 33.62 & 13.25 \\
 &  & 20 & 11.10 & \textbf{24.23} & 150.49 & 30.93 & 9.01 & \textbf{19.73} & 59.61 & 29.85 \\
 &  & 30 & 11.10 & \textbf{47.55} & 2907.95 & 97.16 & 9.01 & \textbf{50.85} & 231.09 & 98.29 \\[2pt]
Llama-3.2-3B & 28 & 10 & 9.02 & \textbf{11.93} & 19.10 & 12.08 & 7.60 & \textbf{9.39} & 13.54 & 9.59 \\
 &  & 20 & 9.02 & \textbf{17.87} & 30.66 & 22.77 & 7.60 & \textbf{13.79} & 21.06 & 17.28 \\
 &  & 30 & 9.02 & \textbf{32.69} & 72.10 & 68.01 & 7.60 & \textbf{28.34} & 44.66 & 46.47 \\[2pt]
Gemma-3-1B & 16 & 10 & 16.48 & \textbf{22.18} & 27.49 & 34.57 & 11.98 & \textbf{15.54} & 17.67 & 19.85 \\
 &  & 20 & 16.48 & \textbf{33.23} & 38.28 & 59.43 & 11.98 & \textbf{23.55} & 25.39 & 39.80 \\
 &  & 30 & 16.48 & \textbf{50.89} & 60.92 & 116.29 & 11.98 & \textbf{40.42} & 49.28 & 110.21 \\[2pt]
Gemma-3-4B & 12 & 10 & 10.79 & \textbf{13.89} & 15.12 & 17.25 & 9.00 & \textbf{12.20} & 13.46 & 14.09 \\
 &  & 20 & 10.79 & \textbf{17.15} & 24.74 & 23.32 & 9.00 & \textbf{16.37} & 26.63 & 20.88 \\
 &  & 30 & 10.79 & \textbf{22.57} & 40.12 & 45.49 & 9.00 & \textbf{23.92} & 43.15 & 44.14 \\[2pt]
Qwen3-1.7B & 24 & 10 & 10.21 & \textbf{13.07} & 16.40 & 14.48 & 7.76 & \textbf{9.88} & 11.87 & 10.70 \\
 &  & 20 & 10.21 & \textbf{18.73} & 26.29 & 21.37 & 7.76 & \textbf{13.74} & 16.57 & 16.10 \\
 &  & 30 & 10.21 & \textbf{35.60} & 57.04 & 45.49 & 7.76 & \textbf{26.42} & 35.50 & 42.63 \\[2pt]
Qwen3-8B & 32 & 10 & 7.71 & \textbf{9.25} & 11.01 & 9.64 & 6.31 & \textbf{7.29} & 8.51 & 7.48 \\
 &  & 20 & 7.71 & \textbf{11.71} & 16.45 & 12.23 & 6.31 & \textbf{9.00} & 10.85 & 9.42 \\
 &  & 30 & 7.71 & \textbf{16.95} & 26.45 & 23.34 & 6.31 & \textbf{13.12} & 15.87 & 14.31 \\
\bottomrule
\end{tabular}
\end{table}

For each model and dataset, we score a block of 8,192 held-out target tokens
preceded by 1,024 context tokens. The PubMed test stream concatenates
abstracts from the held-out article split. To measure MLP latency, we use the same
512-token input from the corresponding stream for all four methods. We randomize
method order in each of five rounds. For each method, we warm up with three passes
per round and time eight more. We record CUDA events at the entry and exit of each
decoder MLP and read the elapsed times after GPU synchronization at the end of the
pass. For the dense MLP time share in Section~\ref{sec:cost-accounting}, we divide
the sum of these intervals by the CUDA-event time for the enclosing forward pass.
Each model contributes 240 dense passes (40 per dataset and savings target).
Table~\ref{tab:mlp-latency-full} gives the per-setting medians used in
Table~\ref{tab:mlp-latency}.

\begin{table}[htbp]
\centering
\footnotesize
\setlength{\tabcolsep}{3pt}
\caption{Median decoder MLP latency (ms) by model, validation-selected group count
$G$, MLP projection savings target, and corpus. Each entry is the median of 40
passes on an RTX 5090, summing MLP times across layers for one 512-token sequence
per pass. Ch. and Grp. denote channel top-$k$ and grouped top-$k$.}
\label{tab:mlp-latency-full}
\begin{tabular}{@{}lrrrrrrrrrr@{}}
\toprule
& & & \multicolumn{4}{c}{WikiText-2} & \multicolumn{4}{c}{PubMed} \\
\cmidrule(lr){4-7}\cmidrule(l){8-11}
Model & $G$ & Save\,\% & Dense & PSSE & Ch. & Grp. & Dense & PSSE & Ch. & Grp. \\
\midrule
Llama-3.2-1B & 12 & 10 & 5.19 & 4.79 & 4.75 & 4.74 & 5.21 & 4.81 & 4.77 & 4.76 \\
 &  & 20 & 5.16 & 4.48 & 4.60 & 4.60 & 5.21 & 4.49 & 4.61 & 4.61 \\
 &  & 30 & 5.17 & 4.10 & 4.32 & 4.32 & 5.23 & 4.11 & 4.33 & 4.32 \\[2pt]
Llama-3.2-3B & 28 & 10 & 12.67 & 10.73 & 10.40 & 10.40 & 12.68 & 10.74 & 10.44 & 10.44 \\
 &  & 20 & 12.68 & 10.12 & 10.12 & 10.12 & 12.69 & 10.13 & 10.13 & 10.13 \\
 &  & 30 & 12.68 & 9.19 & 9.91 & 9.91 & 12.69 & 9.19 & 9.91 & 9.91 \\[2pt]
Gemma-3-1B & 16 & 10 & 5.16 & 4.78 & 4.97 & 4.97 & 5.16 & 4.78 & 4.98 & 4.97 \\
 &  & 20 & 5.15 & 4.29 & 4.31 & 4.32 & 5.17 & 4.30 & 4.33 & 4.33 \\
 &  & 30 & 5.16 & 3.69 & 3.47 & 3.47 & 5.15 & 3.70 & 3.48 & 3.48 \\[2pt]
Gemma-3-4B & 12 & 10 & 15.02 & 13.95 & 14.25 & 14.25 & 15.01 & 13.97 & 14.24 & 14.23 \\
 &  & 20 & 15.03 & 12.75 & 12.86 & 12.86 & 15.04 & 12.77 & 12.88 & 12.88 \\
 &  & 30 & 15.01 & 11.73 & 11.46 & 11.46 & 15.04 & 11.74 & 11.47 & 11.47 \\[2pt]
Qwen3-1.7B & 24 & 10 & 7.81 & 6.86 & 6.67 & 6.67 & 7.81 & 6.87 & 6.67 & 6.67 \\
 &  & 20 & 7.81 & 5.84 & 5.76 & 5.76 & 7.81 & 5.84 & 5.76 & 5.76 \\
 &  & 30 & 7.82 & 5.43 & 5.38 & 5.37 & 7.82 & 5.43 & 5.38 & 5.38 \\[2pt]
Qwen3-8B & 32 & 10 & 28.66 & 25.98 & 26.69 & 26.68 & 29.07 & 26.28 & 26.86 & 26.87 \\
 &  & 20 & 29.03 & 22.78 & 22.68 & 22.69 & 29.21 & 22.96 & 22.86 & 22.87 \\
 &  & 30 & 29.06 & 20.56 & 21.31 & 21.32 & 29.21 & 20.63 & 21.36 & 21.37 \\
\bottomrule
\end{tabular}
\end{table}

\begin{table}[!htbp]
\centering
\small
\caption{Layer-width allocation ablation at matched projection FLOPs. Each pair gives
prefix-based / expected-active perplexity; bold marks the lower value. The expected-active
target is proportional to $\sum_g |I_g|\pi_g$ within each layer, using the same group
probabilities as the prefix. Both allocators use the same feasible width choices and
PSSE-selected $G$; each selects channel or group retention by the same validation procedure.}
\label{tab:width-allocation}
\begin{tabular}{@{}llrrr@{}}
\toprule
& & \multicolumn{3}{c}{Target MLP projection FLOP savings} \\
\cmidrule(l){3-5}
Model & Dataset & 10\% & 20\% & 30\% \\
\midrule
Llama-3.2-1B & WikiText-2 & \textbf{15.12} / 15.75 & \textbf{24.23} / 25.52 & \textbf{47.55} / 58.13 \\
 & PubMed & \textbf{12.31} / 13.12 & \textbf{19.73} / 23.33 & 50.85 / \textbf{49.06} \\[2pt]
Llama-3.2-3B & WikiText-2 & \textbf{11.93} / 13.03 & \textbf{17.87} / 22.18 & \textbf{32.69} / 44.60 \\
 & PubMed & \textbf{9.39} / 10.36 & \textbf{13.79} / 17.47 & \textbf{28.34} / 41.17 \\[2pt]
Gemma-3-1B & WikiText-2 & \textbf{22.18} / 23.00 & \textbf{33.23} / 34.67 & 50.89 / \textbf{50.56} \\
 & PubMed & 15.54 / \textbf{15.44} & 23.55 / \textbf{23.39} & 40.42 / \textbf{37.55} \\[2pt]
Gemma-3-4B & WikiText-2 & \textbf{13.89} / 14.85 & \textbf{17.15} / 20.63 & \textbf{22.57} / 24.97 \\
 & PubMed & \textbf{12.20} / 13.89 & \textbf{16.37} / 22.38 & \textbf{23.92} / 27.72 \\[2pt]
Qwen3-1.7B & WikiText-2 & \textbf{13.07} / 17.65 & \textbf{18.73} / 24.34 & \textbf{35.60} / 38.08 \\
 & PubMed & \textbf{9.88} / 12.60 & \textbf{13.74} / 17.55 & \textbf{26.42} / 33.84 \\[2pt]
Qwen3-8B & WikiText-2 & \textbf{9.25} / 10.61 & \textbf{11.71} / 15.51 & \textbf{16.95} / 24.79 \\
 & PubMed & \textbf{7.29} / 8.09 & \textbf{9.00} / 10.14 & \textbf{13.12} / 13.93 \\
\bottomrule
\end{tabular}
\end{table}

\FloatBarrier

\section{Extended Related Work}
\label{sec:extended-related-work}

\subsection{Classical State Enumeration}
Our work is inspired by classical \emph{state enumeration} methods developed in reliability theory for analyzing large systems with unreliable components. In this literature, a system is modeled as a collection of components whose joint failure configurations define an exponentially large state space. Algorithms such as ORDER and ORDER-II enumerate system states in descending order of probability mass and approximate expected system performance by evaluating only the most probable states, terminating once sufficient probability mass has been covered \citep{li1984performance,lam1986improved}. These methods yield accurate approximations with strong empirical convergence properties while avoiding exhaustive enumeration.

\subsection{Input-Dependent Sparsity}
More recent methods induce input-dependent sparsity at inference time, including early-exit architectures and adaptive token pruning in transformers \citep{scardapane2024conditional}. DynamicViT, for example, removes redundant image patches in vision transformers \citep{rao2021dynamicvit}. Related work on efficient pretraining drops less informative tokens from intermediate BERT layers and restores the full sequence at the final layer \citep{hou2022token}. These approaches typically rely on fixed hyperparameters such as a target sparsity ratio or confidence threshold.

\subsection{Sparse and Approximate Attention}
For a sequence of length $n$, sparse attention can reduce the $O(n^2)$ cost of
full attention by restricting which tokens attend to one another. Block-local and
strided patterns reduce the complexity to $O(n\sqrt{n})$~\citep{child2019generating}.
Longformer and BigBird combine local windows with global attention on selected
tokens~\citep{beltagy2020longformer,zaheer2020big}. Reformer uses locality-sensitive
hashing to group similar queries and keys, restricting the pairs it compares on
long sequences~\citep{kitaev2020reformer}. The top-$k$ attention method scores keys for each
query and retains the $k$ largest scores~\citep{gupta2021memory}.

Other methods approximate the attention computation directly. Linformer
projects keys and values to a lower dimension to obtain a low-rank approximation
of the attention matrix~\citep{wang2020linformer}. Performer uses random features
to approximate the softmax kernel~\citep{choromanski2021rethinking}. PSSE orders
complete configurations of included and excluded factors by joint probability
and truncates the sum when their cumulative mass reaches the target.

\subsection{Analysis of Effective Model Usage}
Removing individual attention heads can reveal redundancy and specialization
in Transformers~\citep{michel2019sixteen,voita2019analyzing}. Network dissection
relates the activations of individual units to semantic concepts~\citep{bau2017network}.
For decoder MLPs, PSSE measures how many joint configurations account for a given
share of the probability mass. The distinct channels represented in those
configurations determine how we allocate the pruning budget across layers.

\end{document}